\documentclass[11pt, letterpaper]{article} 
\usepackage{fullpage}
\usepackage{graphicx}
\usepackage{amsmath}
\usepackage{booktabs}
\usepackage{amssymb}
\usepackage{multicol}
\usepackage{mathtools}
\usepackage{blkarray}
\newcommand{\matindex}[1]{\mbox{\scriptsize#1}}
\usepackage{hyperref}
\usepackage{url}
\usepackage{lmodern}
\usepackage[T1]{fontenc}

\usepackage{parskip}
\newcommand{\xbf}{\mathbf{x}}
\newcommand{\xdbf}{\mathbf{x_D}}
\newcommand{\xibf}{\mathbf{x_I}}

\newcommand{\Xbf}{\mathbf{X}}
\newcommand{\Anbf}{\mathbf{A_n}}
\newcommand{\Ar}{\mathbf{A_r}}
\newcommand{\Abf}{\mathbf{A}}
\newcommand{\Arht}{\mathbf{\hat{A}_r}}

\newcommand{\Adrht}{\mathbf{\hat{A}_{D,r}}}
\newcommand{\Airht}{\mathbf{\hat{A}_{I,r}}}

\newcommand{\Adr}{\mathbf{A_{D,r}}}
\newcommand{\Air}{\mathbf{A_{I,r}}}

\newcommand{\Ad}{\mathbf{A_D}}
\newcommand{\Ai}{\mathbf{A_I}}

\newcommand{\Buf}{\mathbf{B_f}}
\newcommand{\Cuf}{\mathbf{C_f}}

\begin{document}

\title{\textbf{Network Topology Identification using PCA and its Graph Theoretic Interpretations}}
\author{
	Aravind Rajeswaran \hspace{10pt} Shankar Narasimhan\footnote{Corresponding author, Email: naras@iitm.ac.in} \\[10pt]
	\normalsize Systems $\&$ Control Group, Department of Chemical Engineering, \\ 
	\normalsize Indian Institute of Technology Madras, Chennai - 600036, India
}
\date{\normalsize \today}
\maketitle

\begin{abstract}
We solve the problem of identifying (reconstructing) network topology from steady state network measurements. Concretely, given only a data matrix $\Xbf$ where the $X_{ij}$ entry corresponds to flow in edge~$i$ in steady-state~$j$, we wish to find a network structure for which flow conservation is obeyed at all the nodes.
This models many network problems involving conserved quantities including water, power, metabolic networks, and epidemiology.
We show that identification is equivalent to learning a model $\Anbf$ which captures the approximate linear relationships between the different variables comprising $\Xbf$ (i.e. of the form $\Anbf \Xbf \approx \mathbf{0}$) such that $\Anbf$ is full rank (highest possible) and consistent with a network node-edge incidence structure.
We solve this problem through a sequence of steps like estimating approximate linear relationships using Principal Component Analysis, obtaining fundamental cut-sets from these approximate relationships, and graph realization from f-cut-sets (or equivalently f-circuits). 
Each step and the overall process is polynomial time. The method is illustrated by identifying topology of a water distribution network. We also study the extent of identifiability from steady-state data. \\

\noindent {\bf \em Key words:} Network reconstruction, Low rank approximation, PCA, Graph realization
\end{abstract}

\noindent \textbf{Note:} Initially uploaded on July 2, 2015. Revised version uploaded on Jan 22, 2016. Methods and results are unchanged. Introductory section has been revised, and a more detailed literature survey has been added.
\vspace{-10pt}
\section{Introduction and Problem Definition}
\vspace{-10pt}
Learning or estimating models and relationships from data is ubiquitous and of importance to nearly every scientific discipline. In the unsupervised learning setup~\cite{MLbook}, the objective is to find hidden structure in unlabeled data. In other words, there is no explicit mapping from one set of variables to another which we are trying to learn. Instead the objective is to find some low dimensional manifold which captures nearly all variability in the data. However, the models learnt this way are often not interpretable~\cite{SPCA06}. Interpretability sheds light on the mechanics of data generating process, which can be used in subsequent exercises like design or decision making.

A major characteristic of interpretable models are their structures. By looking at the structure, it might be possible to interpret or attribute physical meaning to different groups of coefficients or terms. Hence it is required to either learn structured models directly, or transform a learnt (black-box) model into the desired structure. In most cases, we do not possess the exact parametrized structural form of the model (if that were the case, the task would reduce to simple parameter estimation). Rather, what we possess are broad defining characteristics of the structures, like non-negativity in NMF~\cite{NMF} and sparsity in compressed sensing~\cite{Donoho} or SPCA~\cite{SPCA06}.

In this paper, we look at a particular problem in the broad class of learning interpretable models. Our objective is to identify (or learn) a network topology that connects the different variables in the given data. The work is motivated by mathematical curiosity and applications which require identifying underlying network topology like smart grids~\cite{Arya, Jayadev}, metabolic networks~\cite{Maha04}, and water networks~\cite{Naras15}. The method developed is inspired by Principal Component Analysis (PCA)~\cite{Jolliffe}, its various spin-offs~\cite{MLPCA, Naras15, SPCA06}, and their graph theoretic interpretations.

As input, we are given only a data matrix $\Xbf$ (size $m \times d$), the rows of which describe different variables $x_1, \ldots x_m$ and, columns correspond to noisy measurements under different scenarios or configurations. In other words, the entry $X_{ij}$ corresponds to variable $x_i$ is the $j^{\text{th}}$ configuration. 
There are $m$ variables and we possess full set of measurements (i.e.~each variable measured) corresponding to $d$ configurations.
We seek to find a relationship between the variables, that are common across the different configurations. In this paper, we restrict our attention to approximate linear relationships (ALRs). In other words, we seek a ``model'' $\Anbf$ (size $n \times m$) that captures the ALRs of the form $\Anbf \Xbf \approx 0$ and is also consistent with a network incidence structure (emphasized by the subscript). We also require that $\Anbf$ have full rank (highest possible) to ensure that we get the network with maximum number of nodes. Also note that we are not given the number of equations (or equivalently the rank of $\Xbf$ or number of nodes) and hence we try and capture all possible predictability in the data. The structural properties and requirements of a network incidence structure are reviewed in Section~4. 

Alternatively, we wish to solve the following optimization problem:
\begin{equation}
\begin{aligned}
& \underset{\mathbf{\Anbf, \hat{\Xbf}}}{\text{minimize}}
& & \sum_{i=1}^{m} \sum_{j=1}^{d} \left( \frac{X_{ij}-\hat{X}_{ij}}{\sigma_{ij}}  \right)^2
\\
& \text{subject to}
& & \Anbf \hat{\Xbf} = \mathbf{0}_{n \times d} \\
& {}
& & \Anbf \in \Omega (n,m) \\
& {}
& & \text{rank}(\Anbf) = n
\end{aligned}
\end{equation}
where $\Omega(n,m)$ is the set of all allowable incidence matrices or configurations with $n$ nodes and $m$ edges (including dangling edges, see Section 4). Here, $n$ must be specified for the optimization problem, and we wish to pick the maximum possible $n$ for which a solution exists.

We interpret the learnt model as follows: there is a flow network whose edge flows are the variables $x_1, x_2 \ldots x_m$ and flow conservation is obeyed at the nodes. Some examples are flow networks (say water) where we posses edge flow measurements by some means, and wish to infer either partial or total connectivity of the network. Similarly in case of power grids, we may possess data on amount of power transmitted through edges but may not know how the edges are connected (i.e. the node-edge incidence structure). In other words, the problem is relevant to any case which involves edge flows, using which we try to identify or reconstruct the network topology. The network can describe an actual physical network as in the earlier examples, or may simply describe abstract relationships between variables in a convenient or concise manner.

Needless to say, identifying network topology can provide many insights about the system. For example, degree distribution reveals the presence of any central or hubs in the network; clustering coefficient can provide insight about redundancy and hence robustness to failures; and betweenness centrality can provide an insight about bottlenecks and flow traffic in the network, thereby giving us a good understanding of which variables are difficult to control. Thus identifying the network structure will enable us to understand and manipulate it better.

In this paper, we provide a polynomial time algorithm for network topology identification. We also answer closely related and important questions like (a)~Under what conditions is network identifiable? (b) How many different configurations (i.e. data) are required? (c) Is the realized network unique? and (d) Are there any limitations when working with data of this steady-state form (no temporal correlation between different configurations)? 

\vspace{-10pt}
\section{Connections to Related Work} 
\vspace*{-10pt}
Related work has occurred on two separate fronts - obtaining effective ALRs and network reconstruction. Principal Component Analysis (PCA) is a widely used technique for two broad purposes: dimensionality reduction and estimating ALRs. We provide a brief review of PCA in Section~5 which is required for describing our method. The simpler problem of obtaining ALRs is well studied and considered solved under mild assumptions~\cite{Jolliffe, Naras15}. However it is also well known that interpreting these relationships and obtaining physical insights is a hard task~\cite{SPCA06}. This is largely because of a {\it rotation ambiguity} which is a result of choosing an orthogonal basis set (chosen adaptively based on data) to represent the ALRs, which may not posses a direct physical interpretation.

To provide physical meaning to the discovered ALRs, different methods have been proposed. One of the most successful among these is based on the notion that a sparse representation is likely to provide better physical interpretation. Though this is intuitive and has been successful, it does not solve all problems. For instance, in our problem, though the model we wish to recover (i.e. $\Anbf$) is sparse - it also possess other important structural properties. Thus a sparse representation by itself does not necessarily solve the problem. The structure of sparsity pattern is also of great importance. Some problems in the realm of structured sparsity have also been studied recently~\cite{SPCA07, Huang11} where the idea is to incorporate appropriate structural constraints in an optimization framework. However, some of these constraints can make the problem combinatorial, as in our case, which is generally handled through relaxations. In this work, we however take a different approach. We try and transform the estimated model into a structured sparsity form instead of directly estimating the structured model. In our problem, this provides advantages like reduced computational effort, and more importantly lends itself to good theoretical understanding. For example, some of the questions raised at the end of Section~1 cannot be answered easily if we just use an optimization problem with structural constraints. For instance, it is difficult to characterize multiplicity of solutions or the connections between them.

On the other front, network reconstruction is generally studied in the context of dynamical systems and time-series data~\cite{AKT1, Yuan}. Relevant examples include identifying plant topology or connectivity from time series measurements~\cite{AKT1}, reconstruction of chemical reaction networks from data~\cite{Yuan} etc. These exercises involve fitting an appropriate time series model (like Vector Auto Regressive models), and inferring connectivity from the coefficients of fitted models. However, obtaining the same connectivity information from steady state data is a difficult problem. This is again due to the rotation ambiguity. To the best of the authors knowledge, there has not been any work involving network reconstruction from edge flow measurements. This is likely because for a large number of applications, nodal throughput measurements are likely easier than measuring edge flows~\cite{Arya}. However, the problem of reconstructing networks from edge measurements is important for theoretical completeness, and also serves as an example to illustrate that it may be possible to incorporate structure as a post processing step as opposed to incorporating structural constraints in the estimation step. Also, edge measurements seem appropriate in newer applications like network tomopraphy~\cite{Castro04} and we expect the proposed method to find more applications in the near future.

\section{Method overview and example}
\vspace*{-10pt}
We provide a concrete example to illustrate the problem we are trying to solve. Consider the flow network shown in Fig.~1. We have depicted representations which allows for {\it dangling edges} (incident on only one node) as well as closed representations of the same (connect free end of all dangling edges to one {\it environment} node) which is a standard practice in nearly all problems involving flow networks. There are a total of 6 flow variables ($x_1 \ldots x_6$). For simplicity, assume noiseless measurements of all 6 edges. If flow conservation is active in the nodes, we can represent the relationships between $x_1, x_2 \ldots x_6$ as shown below.
$$ \mathbf{A_n x = 0} $$
$$ \Anbf = 
\begin{bmatrix}
1 & 1 & -1 & 0 & 0 & 0 \\
0 & 0 & 1 & -1 & 0 & 0 \\
0 & 0 & 0 & 1 & -1 & -1 \\
0 & -1 & 0 & 0 & 0 & 1 \\
\end{bmatrix}
$$
$$ \mathbf{x} =
\begin{bmatrix}
x_1 & x_2 & x_3 & x_4 & x_5 & x_6
\end{bmatrix}^T $$

\begin{figure}[t!]
\begin{center}
\includegraphics[width=0.5\textwidth]{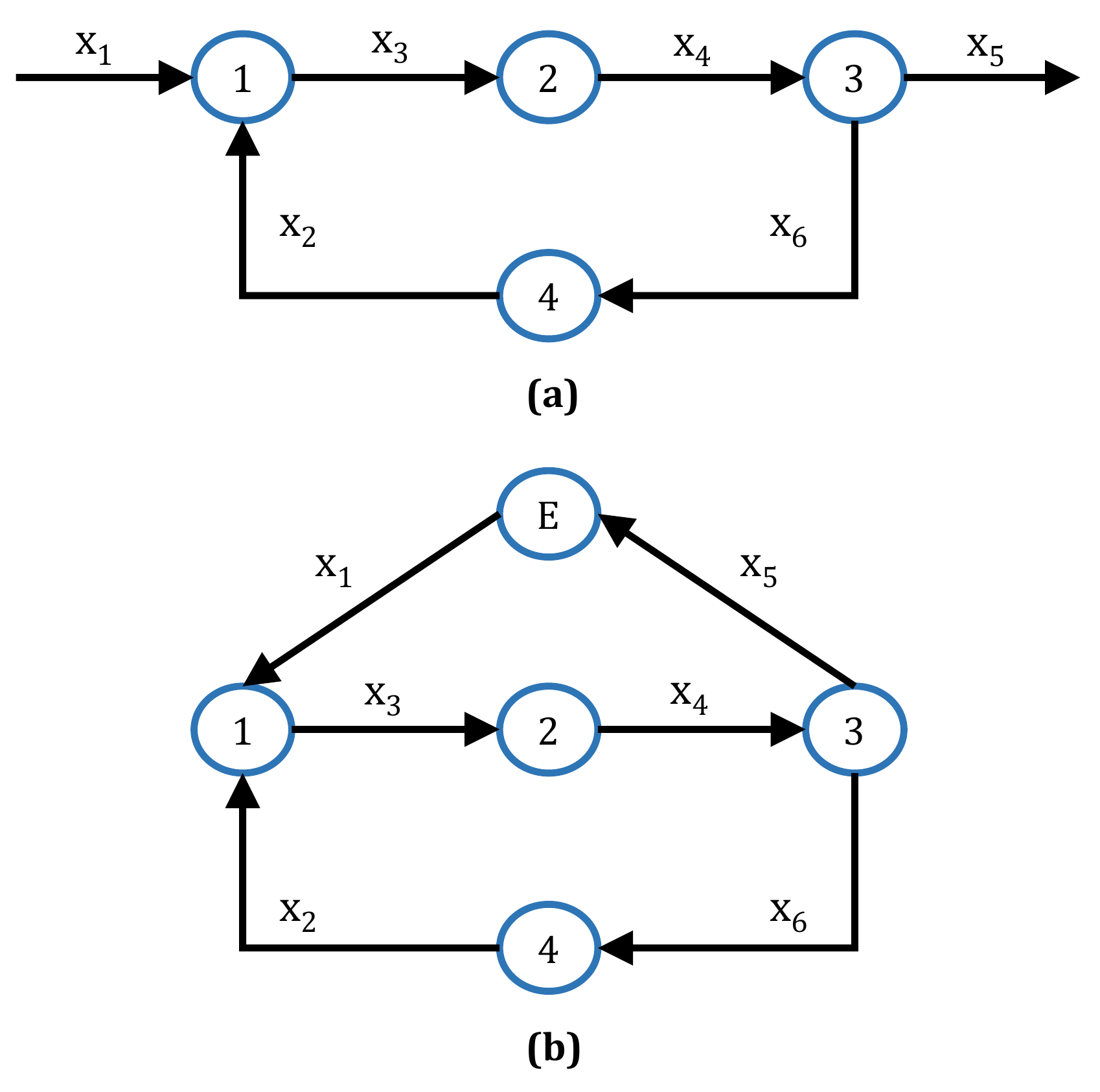}
\caption{An illustrative example of flow network in (a) reduced form and (b) closed form}
\end{center}
\label{Flow_Network}
\end{figure}

Given the flow data (i.e. $\Xbf$) in Table~1, the objective is to use this data alone for reconstructing or inferring the underlying network topology. In this example, we have measurements corresponding to 7 different steady states (i.e. $d = 7$).

\begin{table}
\caption{Flow data corresponding to network in Fig.~1}
\begin{center}
\begin{tabular}{@{}crrrrrrr@{}}
\toprule
Variable & \multicolumn{7}{c}{Measurements} \\ \midrule
$x_1$        & 1  & 1  & 1  & 2  & 2  & 2  & 3  \\
$x_2$        & 1  & 2  & 3  & 1  & 2  & 3  & 1  \\
$x_3$        & 2  & 3  & 4  & 3  & 4  & 5  & 4  \\
$x_4$        & 2  & 3  & 4  & 3  & 4  & 5  & 4  \\
$x_5$        & 1  & 1  & 1  & 2  & 2  & 2  & 3  \\
$x_6$        & 1  & 2  & 3  & 1  & 2  & 3  & 1  \\ \bottomrule
\end{tabular}
\end{center}
\end{table}

\begin{figure}[t!]
\label{Flow_Sheet}
\begin{center}
\includegraphics[width=0.45\textwidth]{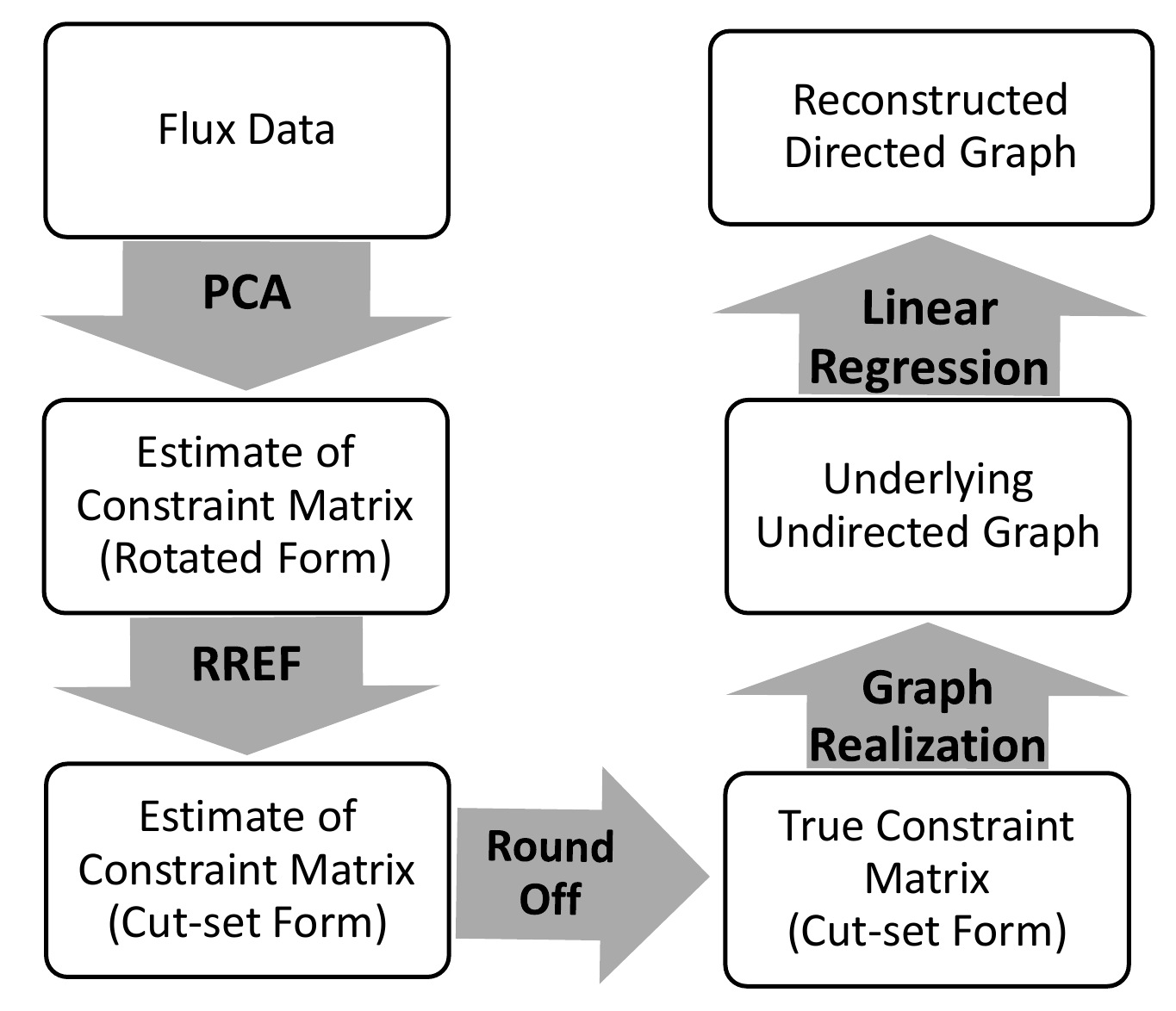} \\
\caption{Sequence of steps in the proposed method.}
\end{center}
\end{figure}

The method we propose involves multiple steps. First, we obtain the ALRs (or model) using Principal Component Analysis (PCA). This model however cannot be interpreted directly as a network structure. The crux of our contribution involves a method to transform and interpret the discovered approximate relationships as a network structure. The following steps will transform these set of ALRs to a different but equivalent set of linear relationships that are consistent with network models, thereby enabling us to obtain the network topology just by looking at these transformed set of equations. We first obtain the fundamental cut-sets of the network from the PCA estimate. This transformation restores network identifiability. Next, we obtain the underlying undirected graph structure given a spanning tree (minimal) and associated cut-sets, which are obtained in the previous step. This is possible within 2-isomorphism, and the problem is generally called graph realization from cut-sets. Finally, the direction of each edge is obtained by performing linear regression on the structured model obtained using the structure of underlying undirected graph. The process is summarized in Fig. 2. The overall method is polynomial time, since each step of the method is polynomial in number of arithmetic operations.


\section{Review of algebraic graph theory} 

A graph $\mathbf{G} = (\mathbf{N},\mathbf{E})$ contains a set of nodes $(\mathbf{N})$ and associated edges $(\mathbf{E})$. The number of nodes will be denoted by $n$ and the number of edges by $m$ throughout this paper. Unless explicitly stated, we consider connected graphs with no self loops. The incidence matrix $\Anbf$ (size $n \times m$) describes how the edges are incident on nodes. The rows represent nodes and the columns represent edges.
$$ A_{n(ik)} = 
\begin{cases}
+1  & \text{if edge k enters node } i \\
-1 & \text{if edge k leaves node } i \\
0 & \text{if edge k is not incident on node } i
\end{cases}
$$
Due to its particular structure, this matrix is most useful for network reconstruction. It is trivial to verify that the steady-state flow conservation equations can be written as
$$ \sum_{k} A_{n(ik)} x_{k} =  0 \hspace{5pt} \forall i \in \mathbf{N} $$
In other words, the flows are constrained to lie in the nullspace of $\Anbf$. Clearly $\Anbf$ encodes the entire graph structure, and hence estimating $\Anbf$ corresponds to obtaining the underlying network representation which we seek. Each column of $\Anbf$ will have exactly two non-zero entries (one $+1$ and one $-1$) since an edge can leave only one node and enter another. Hence the matrix is sparse and given this information, it is also easy to see that the rank of $\mathbf{A}$ is $n-1$ when working with closed and connected networks~\cite{Deo}, i.e. each edge must originate at a node and terminate at a node.

A closely related form of this matrix is the reduced incidence matrix, which is useful when dealing with dangling edges (edges incident on only one node like $x_1$ in Fig.~1). Similar representations are used for example in metabolic networks when dealing with external fluxes. In such cases these dangling edges are all assumed to be connected to one {\it environment node}, but not explicitly represented. The columns corresponding to dangling edges in $\Anbf$ will have only one non-zero entry (either $+1$ or $-1$). The dimensions of reduced incidence matrix are $n \times m$ where $n$ is the number of nodes in the system we are considering (excluding environment node). Hence it is still one rank lesser than the {\it true} graph which also contains the environment. We denote both forms of the incidence matrix using $\Anbf$ and it is easy to identify based on context.

Other useful matrices for this paper are the fundamental circuit $(\Buf)$ and fundamental cut-set~$(\Cuf)$ matrices. If we consider any spanning tree, with respect to this tree, it is possible to construct a set of circuits and cut-sets which are said to be fundamental, since all other circuits and cut-sets can be represented as their ring sums. The definitions of these matrices again are analogous to the incidence matrix. These matrices set up the relationships between the edges and fundamental circuits or cut-sets.
$$ B_{ij} = 
\begin{cases}
+1 & \text{edge j is in circuit i, same direction} \\
-1 & \text{edge j is in circuit i, opposite direction} \\
0 & \text{otherwise}
\end{cases}
$$    
and similarly for the cut-set matrix,
$$ C_{ij} = 
\begin{cases}
+1 & \text{edge j is in cut-set i, same direction} \\
-1 & \text{edge j is in cut-set i, opposite direction} \\
0 & \text{otherwise}
\end{cases}
$$ 
Further, since these fundamental matrices are written with respect to a spanning tree in mind, a graph can have multiple fundamental circuit or cut-set matrices (by choosing a different tree). It is also true that multiple graphs can have the same $\Buf$ and $\Cuf$ matrices even when they are not isomorphic. Such examples are called 2-isomorphic graphs~\cite{Deo}, and characterize the multiplicity (in fact, exactly) of solution in our approach. This aspect will be addressed in detail in later sections.

\noindent {\bf \em Definition: } Two graphs $G_1$ and $G_2$ are said to be 2-isomorphic if they share circuit (or equivalently cut-set) correspondence. This means that there is a one to one correspondence between edges such that, whenever a set of edges in $G_1$ forms a circuit (or cut-set), the corresponding edges in $G_2$ also form a circuit (or cut-set).

\noindent {\bf \em Fact: } Two graphs $G_1$ and $G_2$ are 2-isomorphic if and only if there exists trees $T_1$ and $T_2$ (associated with $G_1$ and $G_2$ respectively) such that the f-circuit and f-cut-set matrix obtained with respect to $T_1$ and $T_2$ are element wise identical subject to row and column permutations.

\noindent {\bf \em Proof: } Follows from definition. The row and column permutations correspond to 1-isomorphic operations or simply renaming nodes and edges. \hspace{10pt} {\em end of proof.}

We also note that f-circuit and f-cut-set matrices can be decomposed (after a possible column permutation) as:
$$ \Buf = [\mathbf{B_t} \mid \mathbf{I_{\mu}}] $$
$$ \Cuf = [\mathbf{I_{n-1}} \mid \mathbf{C_c}] $$
where $\mu = m - (n-1)$. The size of $\Buf$ is $(\mu \times m)$ and that of $\Cuf$ is $(n-1 \times m)$. A standard result is that $\mathbf{B_t} = - \mathbf{C_c^T}$ and that $\mathbf{C_f B_f^T = 0}$ (mod 2). Hence, given one the other is trivial to find, and both the cut-set and circuit matrices describe a graph within 2-isomorphism. We refer the interested readers to Chapter~7 of Narsingh Deo's book~\cite{Deo} for a detailed discussion.


\section{Review of PCA}
\vspace*{-10pt}
PCA and Singular Value Decomposition (SVD) are among the most widely used multivariate statistical techniques. PCA or SVD provides the best low-rank approximation to a given matrix under homoscedastic Gaussian additive noise assumption. Equivalently PCA provides the best ALRs between a set of variables under the previous error characteristics~\cite{Jolliffe}. 

\noindent Consider the scaled co-variance matrix corresponding to the input data matrix.
$$\mathbf{S_x =} \frac{1}{d} \mathbf{X X^T} $$
The data matrix can also be decomposed using SVD in a sum of product form as:
\begin{equation}
\text{SVD:} \hspace{10pt} \frac{1}{\sqrt{d}} \mathbf{X} = \mathbf{U_1 S_1 V_1^T + U_2 S_2 V_2^T}
\end{equation}
where $\mathbf{U_1}$ are the set of orthonormal eigenvectors corresponding to the $p$ largest eigenvalues of $\mathbf{S_x}$ while $\mathbf{U_2}$ are the set of orthonormal eigenvectors corresponding to the remaining $(m-p)$ smallest eigenvalues values of $\mathbf{S_x}$. The matrices $\mathbf{S_1}$ and $\mathbf{S_2}$ are diagonal, whose entries are the square root of the eigenvalues of $\mathbf{S_x}$. It is assumed that the singular values and corresponding principal components are sorted in decreasing order. To perform the decomposition, we require $\mathcal{O}(k_1 m^2 d + k_2 d^3)$ operations~\cite{Golub}.

If the objective is to reduce dimensionality (or obtain a low rank approximation of $\mathbf{X}$), then it suffices to store only the hyperplane on which most of the data resides $(\mathbf{U_1})$ and the projection of data onto this hyperplane. Since the equation of the hyperplane and coordinates of the data points on this hyperplane (projections) are known, the original data can be recovered up to the level of hyperplane. It is assumed that the variations which deviated the data from this hyperplane are due to noise, and hence dimensionality reduction and filtering have been performed simultaneously. The only task that is left is to determine the number of PCs to be retained $(p)$. In absence of any additional information other than data, $p$ is often heuristically chosen based on criteria like percentage variance explained or scree plot~\cite{Jolliffe}. More principled methods for determining $p$ also exist, and use knowledge of error covariances and an iterative approach (see \cite{Naras15} for e.g.).

The same idea can be extended for obtaining ALRs as well. In model identification, we are interested in finding the sub-space on which the data is constrained to lie, and this sub-space is the set of feasible or admissible region allowed by the model. To characterize this sub-space, we need to find a set of basis vectors orthogonal to this sub-space. This is readily obtained from the $(m-p)$ non-retained PCs in case of dimensionality reduction. Hence, the model obtained is given by: 
\begin{equation}
\mathbf{ U_2^T X \approx 0 }
\end{equation}

\subsection{Rotation Ambiguity}
\vspace{-10pt}
We now formally characterize the rotation ambiguity. If we pre-multiply $\mathbf{U_2}$ by any rotation matrix $\mathbf{M}$ we have:
$$\mathbf{ M U_2^T x = M 0 = 0 }$$
Hence, the best that can be obtained from data is the row-space for the true constraint matrix, which in this case is the directed version of incidence matrix, which is what we wish to recover. If we do not impose any structural restrictions, the best that we can get is a basis, which in case of PCA is an orthogonal basis set chosen adaptively based on data input. 

\noindent A very useful alternate interpretation is through optimization. Consider the following optimization set-up for finding the best approximate relationships:
\begin{equation}
\begin{aligned}
& \underset{\mathbf{\Abf, \hat{\Xbf}}}{\text{min.}}
& & \sum_{i} \sum_{j} \left( \frac{X_{ij}-\hat{X}_{ij}}{\sigma_{ij}}  \right)^2
\\
& \text{s.t.}
& & \Abf \hat{\Xbf} = \mathbf{0} \\
& {}
& & \Abf \Abf^T = \mathbf{I}_{p \times p}
\end{aligned}
\end{equation}
The solution to this problem under the assumption that $\sigma_{ij} = \sigma$ is given by PCA with $\Abf = \mathbf{U_2^T}$. Notice that the second constraint is important and is the source of rotation ambiguity. Without this constraint, the problem becomes ill-posed with the solution $\Abf = \mathbf{0}_{p \times m}$ and $\hat{\Xbf} = \Xbf$. In other words, in the absence of any other information about $\Abf$, we can only recover a basis for the approximate relationships and not the relationships themselves.

If the error-covariance matrix is known to be homoscedastic, i.e. of the form $\mathbf{\Sigma_e} = \sigma^2 \mathbf{I}$, an approximate linear model can be identified by performing SVD of the data matrix like in equation~(1) and obtain the model estimate as $\Arht = \mathbf{U_2}$. In case of heteroscedastic error covariance, if $\mathbf{\Sigma_e}$ is known, the Maximum Likelihood estimate of the linear model can be obtained using MLPCA~\cite{MLPCA, Naras08}. If $\mathbf{\Sigma_e}$ is not known, it was shown by Narasimhan and Shah~\cite{Naras08, Naras15} that it is still possible to recover both the model as well as $\mathbf{\Sigma_e}$ by using a method called IPCA.



\section{Proposed Method}
\vspace{-10pt}
{\bf Notation: } Let us denote the true incidence matrix, obeying the structural requirements, by $\Anbf$. The objective is to recover this from a rotated form, which we denote as $\Ar$. The first step is to get an estimate of $\Ar$, denoted by $\Arht$ using PCA.

\noindent Consider the optimization problem proposed in (1). Our approach would involve finding the solution to a relaxed version of this problem, following which graph theoretical approaches would be used to obtain solutions consistent with the original constraints. The original optimization problem was:
\begin{equation*}
\begin{aligned}
& \underset{\mathbf{\Anbf, \hat{\Xbf}}}{\text{minimize}}
& & \sum_{i=1}^{m} \sum_{j=1}^{d} \left( \frac{X_{ij}-\hat{X}_{ij}}{\sigma_{ij}}  \right)^2
\\
& \text{subject to}
& & \Anbf \hat{\Xbf} = \mathbf{0}_{n \times d} \\
& {}
& & \Anbf \in \Omega (n,m) \\
& {}
& & \text{rank}(\Anbf) = n
\end{aligned}
\end{equation*}
First we remove the second and third constraints, and instead introduce $\Abf \Abf^T = \mathbf{I}_{p \times p}$. This reduces the problem to a PCA set-up for which we have efficient algorithms (Section 5), using which we can find $\Arht$. Following this step, we propose an algorithm to transform $\Arht$ to $\Anbf$, which satisfies the second constraint. We will show that $\Arht$ and $\Anbf$ have the same nullspace, and hence the first constraint is automatically satisfied, and there is no change in the objective function.

\subsection{Graph theoretic interpretation of PCA}
\vspace{-10pt}
From equation (3), an estimate of $\Arht$ is $U_2^T$ and represents a basis for the row space of $\Anbf$. A graph theoretic interpretation of the PCA result will help resolve the rotation ambiguity. Consider a partition of the variables ($\xbf$) into dependent ($\xdbf$) and indipendent ($\xibf$) variables. This partition could be on the basis of process knowledge or can be completely data-driven (will be discussed shortly). Once this partitioning has been achieved, we have:
\begin{equation}
    \Arht \xbf = \Adrht \xdbf + \Airht \xibf  \approx \mathbf{0}
\end{equation}
A successful partitioning ensures that $\Adrht$ is invertible and is numerically stable.
\begin{equation}
    \xdbf \approx \Adrht^{-1} \Airht \xibf = \mathbf{\hat{R}} \xibf
\end{equation}
It is easy to see that $\mathbf{R}$ is unique, and can resolve the rotation ambiguity. Consider the incidence matrix, which is of the form $\Anbf = \mathbf{M} \Ar$ where $\mathbf{M}$ is some invertible matrix describing the rotation and scaling.
\begin{equation}
\begin{split}
\xdbf & = - \Ad^{-1} \Ai \xibf = - \left( \mathbf{M} \Adr \right)^{-1} \left( \mathbf{M} \Air \right) \xibf \\
      & = \Adr^{-1} \mathbf{M}^{-1} \mathbf{M} \Air \ \ \xibf  = \mathbf{R} \xibf
\end{split}
\end{equation}
Thus $\Adrht^{-1} \Airht$ can be compared element wise with $\mathbf{R}$ and this can resolve the rotation ambiguity. This has a graph theoretic interpretation too. The act of partitioning the variables into dependent and independent sets, is exactly equivalent to determining a minimum-spanning tree of the graph, and the matrix $\mathbf{R}$ in fact represents the chords associated with the chosen spanning tree $\mathbf{(x_D)}$.

\subsection{Obtaining f-cutsets from PCA estimate}
\vspace{-10pt}
Since $\Anbf$ and $\Cuf$ have the same row space~\cite{Deo}, given any spanning tree corresponding to variables in $\xdbf$, we have
\begin{equation}
\Cuf \xbf = \mathbf{I} \xdbf + \mathbf{C_c} \xibf = \mathbf{0} 
\end{equation}
Hence, the computed $\mathbf{\hat{R}}$ matrix can be related to the network topology through $\Cuf$ as:
\begin{equation}
\xdbf = \mathbf{ - C_c} \xibf \approx \mathbf{\hat{R}} \xibf
\end{equation}
Since every element of $\mathbf{C_c}$ is from $\lbrace -1, 0, +1 \rbrace$; and since it can be element-wise compared with $\mathbf{\hat{R}}$, we can round-off the entries of $\mathbf{\hat{R}}$ and get an accurate estimate of $\mathbf{C_c}$. For achieving this, we need a method to automatically partition $\xbf$ into $\xdbf$ and $\xibf$. 

\noindent {\bf Reduced Row Ehelon Form (RREF):} A matrix is in RREF iff
\begin{itemize}
\item It is in row echelon form.
\item Every leading coefficient is 1 and is the only nonzero entry in its column.
\end{itemize}

\noindent The RREF of a fat full rank matrix (like the one obtained from PCA) can be computed using an algorithm like Gauss-Jordan method in $\mathcal{O}(m^2d)$ arithmetic operations. Unlike the row echelon form, the RREF of a matrix is unique and does not depend on the algorithm used to compute it. Further, RREF is a unique property of the row space, and hence is constant even if the matrix is rotated and scaled by an invertible matrix. Hence it is straight forward to compute $\mathbf{C_f}$ given $\Anbf$  as:
\begin{equation}
\mathbf{C_f} = \text{rref} \left( \Anbf \right)
\end{equation}
It must be noted however that $\Anbf$ and $\mathbf{C_f}$ might require column permutation for the above equation to hold. But once the RREF is computed, the columns can always be permuted such that the first block of the matrix is Identity. Different permutations correspond to different ways of ordering the edges or equivalently 1-isomorphism. The same idea can be used for the estimated matrices as well. By doing so, we get
\begin{equation}
\mathbf{ \hat{C}_f } = \text{rref} \left( \Arht \right)
\end{equation} 
In practice, we observed that the above computation suffers from numerical instabilities. This can however be mitigated by using both row and column pivoting. Using the same argument as before, $\mathbf{ \hat{C}_f }$ is element-wise comparable with $\mathbf{C_f}$ and hence each entry can be rounded-off to the closest among $\lbrace -1, 0, +1 \rbrace$. Thus, we have been able to resolve the rotation ambiguity and obtain the f-cutsets which have been proven to be correct and unique. An illustrative example of each step in computation is presented in Section~7.

\subsection{Graph Realization Problem}
\vspace{-10pt}
The only remaining step is to identify $\Anbf$ given $\mathbf{C_f}$. The graph 2-isomorphism problem was formally defined in Section~4. The corrolory of the definition can be stated as:

\noindent {\bf \em Fact: } Two graphs are 2-isomorphic if and only if the row space of their incidence matrices are identical, after a possible column permutation (1-isomorphism).

\noindent {\bf \em Proof: } The `if' part of the proof is straightforward. If they share the same row-space, then subject to column permutation (one to one correspondence between edges), they share the same $\mathbf{C_f}$ matrix, since it is a property of the row-space. This means that there is a circuit correspondence, and hence by definition 2-isomorphic. The converse argument is also true. For two graphs to be 2-isomorphic, they should produce the same $\mathbf{C_f}$ matrix subject to column permutation. If this is the case, then their incidence matrices have the same row-space (after the necessary permutation) due to the property of RREF.  \hspace{15pt} {\em end of proof}

From the above fact, it follows that uniqueness of reconstruction cannot be guarenteed by any procedure. As an illustration, consider a network for which the incidence-matrix $\Anbf$ results in $\mathbf{A_n x = 0}$. For any invertible matrix $\mathbf{M}$, the following also holds true: $\mathbf{M A_n x = 0}$. If there exists another network whose incidence matrix $\mathbf{A_n'}$ can be written as $\mathbf{A_n' = MA_n}$ for some $\mathbf{M}$, then it is not possible to distinguish between the two networks based on data alone. However, it is possible to recover at-least one network (if it actually exists). This problem is called the graph realization problem from fundamental cut-sets or circuits, and the realization is possible in almost linear time.

We propose to perform the realization procedure in two steps. In the first step, we identify the structure of the underlying undirected graph. Once this structure is determined, the direction of edges are ascertained by performing linear regression using the structured model. The two step procedure involves:
\begin{enumerate}
    \item The sparsity structure of $\mathbf{C_f}$ is considered. This amounts to disregarding the sign of entries, or the direction of cuts. Algorithms from the graph realization literature can be used to find the underling undirected graph topology. Some representative algorithms are the ones developed by Fujishige\cite{Fuji}, Bixby \& Wagner\cite{Bixby}, Parker\cite{Parker}, and Jianping\cite{Jianping}. These have been developed primarily for undirected graphs and use the fundamental circuit matrix as their inputs, whereas we are working with the f-cut-sets. It is however, very easy to obtain one from the other using the following relationships:
        $$ \mathbf{ B_f = [-C_c^{T} \vert I_{\mu}] } $$
    \item Once the structure of underlying network topology is obtained, a simple linear regression is performed using the structured model to obtain the sign of the coefficients.
\end{enumerate}
This completes the sequence of steps needed to reconstruct the network.


\section{Illustrative Example}
\vspace*{-10pt}
Consider the flow network shown in Fig.~1. We obtain measurements of all the flows labeled 1 through to 6, where each flow measurement is corrupted with some noise. We wish to recover the true connectivity from this data alone, without knowledge of the number of nodes in the network. The actual relationships between the variables are presented in Section~2 along with Fig.~1.

\subsection{Data generating process}
\vspace*{-10pt}
\noindent Data is generated using the following process:
\begin{enumerate}
\item $x_1$ and $x_2$ are chosen as independent variables, so that the rest of the flows can be obtained by solving a set of linear equations. The independent variables are perturbed around a base value (see Table~2).
\item True values of $x_3$ $x_4$ $x_5$ and $x_6$ are computed exactly, and a Gaussian noise is added to each of the six flows, after computing true values for all flows.
\end{enumerate}
The base values for simulation and noise are given in Table~2. In this example, we consider each sensor error to be indipendent and uncorrelated with other sensors. Thus the error covariance matrix is diagonal.

\begin{table}[h!]
\begin{center}
\caption{Base values, standard deviation of fluctuation (SDF), and error (SDE)}
\begin{tabular}{@{}llll@{}}
\toprule
Variable & Base Value & SDF      & SDE  \\ \midrule
$x_1$       & 10         & 1        & 0.1  \\
$x_2$       & 10         & 2        & 0.08 \\
$x_3$       & solved     & computed & 0.15 \\
$x_4$       & solved     & computed & 0.2  \\
$x_5$       & solved     & computed & 0.18 \\
$x_6$       & solved     & computed & 0.1  \\ \bottomrule
\end{tabular}
\end{center}
\end{table}

\subsection{Estimating ALRs using PCA}
\vspace*{-10pt}
When SVD is applied to the data matrix $\left( \frac{1}{\sqrt{d}} \mathbf{X} \right)$ from the example, we obtain the following singular values: 
$$ \begin{bmatrix}
34.96 & 2.25 & 0.43 & 0.40 & 0.34 & 0.3
\end{bmatrix} $$
This clearly indicates that there are 4 singular values close to 0, thereby the data is in fact present in a two dimensional subspace of $\mathbb{R}^6$. In other words the data is constrained to lie in the null space of a $4 \times 6$ matrix.  We get the constraint relationships from:
$$ \mathbf{ \hat{A}_r x = U_2^T x \approx 0 } $$ 
$$ 
\Arht =
\begin{bmatrix}
0.099  & 0.107  & 0.612  & -0.773 & 0.068  & 0.047  \\
-0.318 & 0.213  & -0.315 & -0.173 & 0.806  & 0.275  \\
-0.748 & -0.253 & 0.432  & 0.199  & 0.110  & -0.372 \\
-0.045 & -0.737 & 0.077  & -0.009 & -0.023 & 0.669  
\end{bmatrix} $$

This estimate has been arrived at assuming no knowledge of the true error co-variance matrix, which is representative of the worst-case scenario. Also, we have used SVD instead of the more specialized techniques for heteroscedastic error co-variance to illustrate that the method works remarkably well even when only an approximate model is obtained - either from data alone or partially based on data, and partially based on process knowledge.

Clearly, this estimate ($\Arht$) is different from the true network matrix ($\Anbf$), and an element by element comparison is not possible. However, to check if the two matrices do indeed share the same row-space, Narasimhan and Shah~\cite{Naras08} have proposed two criteria: (i) the subspace angle between the row subspace of the estimated and true constraint matrices, and (ii) the sum of orthogonal distances of the row vectors of the estimated constraint matrix from the subspace defined by the rows of the true constraint matrix. 
\begin{enumerate}
\item[(i)] The subspace angle measures the maximum possible angle between the row sub-spaces spanned by the two matrices. For this example, we have a subspace angle of 0.0091 degrees.
\item[(ii)] Here we project $\Arht$ on to the row space of $\Anbf$, and use Frobenius norm between the projection and $\Arht$ for comparison. For the example, the Frobenius norm was calculated to be $8.21 \times 10^{-5}$.
\end{enumerate}
The above values suggest that the correct row-space was identified by PCA. We now compare the regression ($\mathbf{R}$) matrix to show that it resolves the rotation ambiguity. Let $x_1$ and $x_2$ be the choice for independent variables ($\xibf$). 
\vspace*{-20pt}
\begin{multicols}{2}
$$ \mathbf{ \hat{R} } = 
\begin{bmatrix}
1.007  & 0.995  \\
1.015  & 0.987  \\
1.005  & -0.005 \\
-0.001 & 1.001
\end{bmatrix} $$ \\ \vspace*{-10pt}
$$ \mathbf{R} = 
\begin{bmatrix}
1 & 1 \\
1 & 1 \\
1 & 0 \\
0 & 1
\end{bmatrix} $$ 
\end{multicols}
\noindent Element wise comparison clearly shows us that PCA has identified the correct row space. Small deviations from the true values are due to noise which we added.

\subsection{Identifying cut-sets from PCA estimate}
\vspace*{-10pt}
For the purpose of identifying the cut-sets, we perform the RREF transformation on $\Arht$. The algorithm picked $x_1 \ldots x_4$ to form the spanning tree. The estimated and true cut-set matrices are:
\vspace*{-20pt}
\begin{multicols}{2}
$$ \mathbf{ \hat{C}_f } = \begin{bmatrix}
1 & 0 & 0 & 0 & -0.995 & -0.005 \\
0 & 1 & 0 & 0 & -0.001 & -0.999 \\
0 & 0 & 1 & 0 & -1.003 & -0.999 \\
0 & 0 & 0 & 1 & -1.011 & -0.992
\end{bmatrix} $$ \\ \vspace*{-10pt}
$$ \mathbf{ C_f } = 
\begin{bmatrix}
1 & 0 & 0 & 0 & -1 & 0 \\
0 & 1 & 0 & 0 & 0 & -1 \\
0 & 0 & 1 & 0 & -1 & -1 \\
0 & 0 & 0 & 1 & -1 & -1
\end{bmatrix} $$
\end{multicols}
There is a clear element-wise correspondence between  the two, and hence exact recovery of $\mathbf{C_f}$ from data is possible. Note that if the entries in $\mathbf{ \hat{C}_f }$ have large deviations from acceptable values of $\lbrace -1, 0, +1 \rbrace$, this is strong evidence that the original data generating process was not a network.

\subsection{Graph realization from cut-sets}
\vspace*{-10pt}
After obtaining $\mathbf{\hat{C}_f}$ from data, rounding off, and ignoring signs, we obtain the sparsity pattern of the cut-set ($\mathbf{C_u}$) which can be used to identify the topology of underlying undirected graph.
$$ \mathbf{C_u} = \begin{bmatrix}
1 & 0 & 0 & 0 & 1 & 0 \\
0 & 1 & 0 & 0 & 0 & 1 \\
0 & 0 & 1 & 0 & 1 & 1 \\
0 & 0 & 0 & 1 & 1 & 1
\end{bmatrix} $$
Let $\mathbf{A_u}$ denote the incidence matrix of underlying undirected graph. Each column of $\mathbf{A_u}$ has utmost two non-zero entries, which are equal to $1$. We observe that the above form has more than two non-zero entries in columns 4 and 5. We hence perform the matrix operation:
$ R_4 \leftarrow R_3 + R_4 $ and perform mod~2 arithmetic (i.e. set all elements with value 2 equal to 0) we get:
$$ \mathbf{ A_u^{(1)} } = \begin{bmatrix}
1 & 0 & 0 & 0 & 1 & 0 \\
0 & 1 & 0 & 0 & 0 & 1 \\
0 & 0 & 1 & 0 & 1 & 1 \\
0 & 0 & 1 & 1 & 0 & 0
\end{bmatrix} $$ 
\begin{figure}[b!]
\begin{center}
\includegraphics[width=0.35\textwidth]{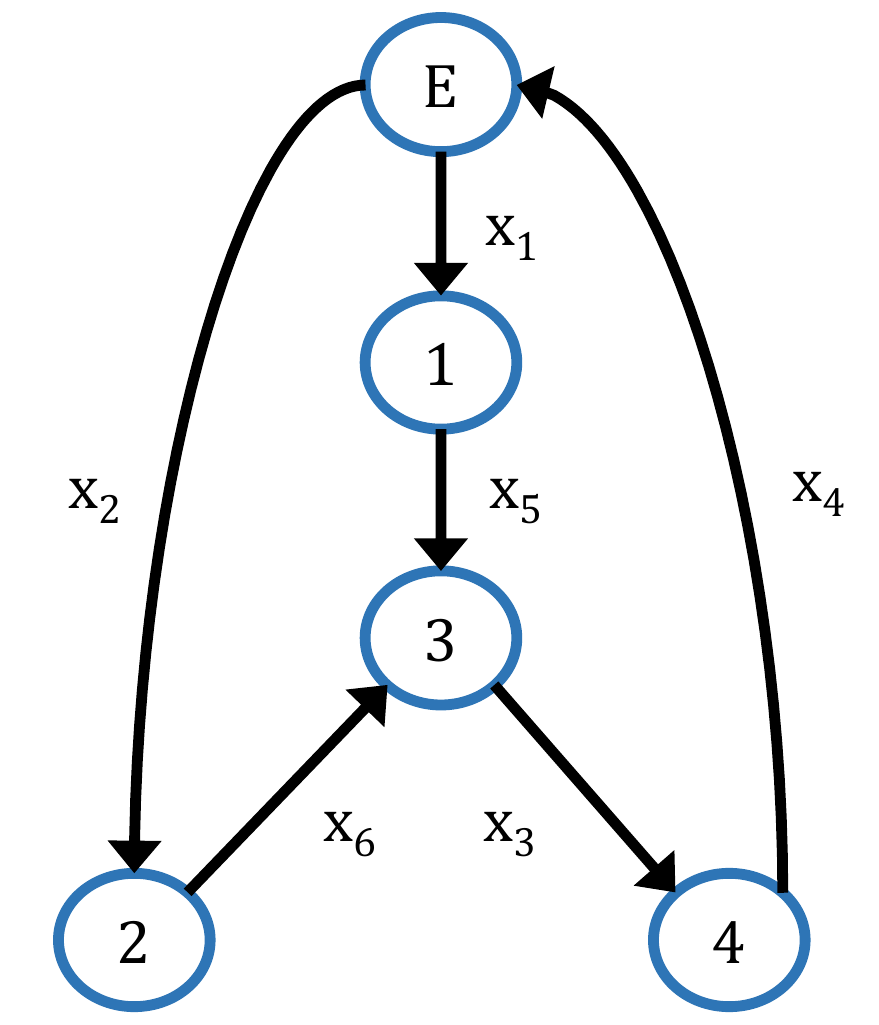} \\
\caption{Reconstructed network}
\end{center}
\end{figure}
\noindent This is in an acceptable network form, and the underlying undirected structure has been recovered. Since the structure has now been obtained, it is trivial to perform linear regression with this particular structure to obtain the direction of edges. This can be done so by estimating a single equation corresponding to only the variables present in each row. The signs of elements in the first row are chosen arbitrarily, and this fixes the signs of all the rows since each column has exactly one positive and one negative entry (including the environment node). If the true data generating process was a network, we expect to get a consistent set of entries without any clashes. The reconstructed network is shown in Fig.~3 and the corresponding reduced indidence matrix is:
$$ \mathbf{ A_\mathrm{est} } = \begin{bmatrix}
1 & 0 & 0 & 0 & -1 & 0 \\
0 & 1 & 0 & 0 & 0 & -1 \\
0 & 0 & -1 & 0 & 1 & 1 \\
0 & 0 & 1 & -1 & 0 & 0
\end{bmatrix} $$ 
\noindent We can see that this graph is different from the original one we started out with. However, the two are completely indistinguishable using flow conservation equations, and both could have generated the data in an equally plausible manner. It is also trivial to verify that the two networks are indeed 2-isomorphic. This can be done by obtaining the RREF of both $\mathbf{A}$ and $\mathbf{A_\mathrm{est}}$, and they would be identical. On the other hand, if we had stacked the variables in an alternate manner (this corresponds to reordering or renaming the edges, i.e. 1-isomorphism) as
$$ \mathbf{x} = \begin{bmatrix}
x_1 & x_3 & x_5 & x_2 & x_4 & x_6
\end{bmatrix} $$
The corresponding undirected cut-set matrix we would obtain after going through the PCA and RREF sequence would be:
$$ \mathbf{ C_u^{(2)} } = 
\begin{blockarray}{cccccc}
\matindex{x1} & \matindex{x3} & \matindex{x5} & \matindex{x2} & \matindex{x4} & \matindex{x6}\\
    \begin{block}{[cccccc]}
	1 & 0 & 0 & 0 & 1 & 1 \\
	0 & 1 & 0 & 0 & 1 & 0 \\
	0 & 0 & 1 & 0 & 1 & 1 \\
	0 & 0 & 0 & 1 & 0 & 1 \\
    \end{block}
\end{blockarray} $$
Performing $R_1 \leftarrow R_1 + R_2$ followed by $R_1 \leftarrow R_1 + R_4$, we have the following incidence matrix.

$$ \mathbf{ A_u^{(2)} } = 
\begin{blockarray}{cccccc}
\matindex{x1} & \matindex{x3} & \matindex{x5} & \matindex{x2} & \matindex{x4} & \matindex{x6}\\
    \begin{block}{[cccccc]}
	1 & 1 & 0 & 1 & 0 & 0 \\
	0 & 1 & 0 & 0 & 1 & 0 \\
	0 & 0 & 1 & 0 & 1 & 1 \\
	0 & 0 & 0 & 1 & 0 & 1 \\
    \end{block}
\end{blockarray} $$
This corresponds to exactly the network we originally used, subject to the same column permutations (i.e. the two graphs are 1-isomorphic). This can be verified by permuting the columns of the original incidence matrix accordingly and comparing element wise with $\mathbf{ A_u^{(2)} }$.

\section{Future Work}
\vspace*{-10pt}
The problem of graph realization from circuits and cut-sets received much attention during the 1980s, but have since remained dormant. Further work on this problem, and more intuitive and efficient algorithms for the same, would prove very useful, given the data deluge today. We have shown in this work that graph realization is an indirect method for estimating network models from data, which is a very important problem today in a number of fields of fields - from economics to biology. We also believe that the proposed method can be extended to a larger set of structured models, in particular the stoichiometry matrix reconstruction. This would provide an alternate method to cross-verify or reconstruct chemical reaction networks using only steady state data.

Also, it may be possible to ascertain the direction of edges after recovering the underlying undirected network topology by just looking at the structure of the signed circuit matrix, and performing a loop traversal. Work on this front could help in reducing one step in the proposed sequence.


\section{Conclusions}
\vspace*{-10pt}
A polynomial time algorithm for network identification from steady state flow data was developed. The method involves a sequence of procedures summarized in Fig.~2. The major components are PCA (or SVD) followed by obtaining the cut-set matrix (in reduced row echelon form) and finally graph realization. The edge directions are ascertained using linear regression. This problem finds application in variety of fields like network topology reconstruction in distribution systems like water and power, and with further work in areas like economics and systems biology. The overall procedure is polynomial time, with the most expensive step being the singular valued decomposition of the data matrix, which can be obtained in $\mathcal{O}(k_1 m^2 d + k_2 d^3)$ where $m$ is the number of edges (or fluxes) and $d$ is the number of distinct steady state data collected~\cite{Golub}. We revisit the questions raised in Section~I and answer them:  
\begin{enumerate}
\item[(i)] Network reconstruction is possible when entries in $\mathbf{\hat{C}_f}$ are not far off from acceptable values of $0$, $1$ or $-1$, and the structure is compatible to a network. This is determined by graph realization algorithms~\cite{Bixby}.
\item[(ii)] We need at least $n$ steady states for reconstruction. Since the value of $n$ may not be known upfront, a safe estimate would be $m$ distinct steady states.
\item[(iii)] The realized network is unique only up-to 2-isomorphism. Any further resolvability is not possible from steady state data alone. Causality cannot be determined as expected - reversing the direction of every edge yields an equally plausible network.
\end{enumerate}

\section*{Acknowledgments}
\noindent The authors would like to thank Dr. Nirav Bhatt, IIT Madras, for his valuable inputs and help in identifying the problem.


\end{document}